\pdfoutput=1

\documentclass[11pt]{article}

\usepackage[final]{acl}
\usepackage{float}
\usepackage{graphicx}
\usepackage{subcaption}
\usepackage{times}
\usepackage{latexsym}

\usepackage[T1]{fontenc}

\usepackage[utf8]{inputenc}

\usepackage{microtype}

\usepackage{inconsolata}

\title{Fine-tuning Pre-trained Named Entity Recognition Models For Indian Languages}

\author{Sankalp Bahad\textsuperscript{1}, Pruthwik            Mishra\textsuperscript{1},
        Karunesh Arora\textsuperscript{2} \\
        , {\bf Rakesh Chandra Balabantaray\textsuperscript{3}},
       {\bf Dipti Misra Sharma\textsuperscript{1}} \and {\bf Parameswari Krishnamurthy\textsuperscript{1}} \\
        LTRC, IIIT Hyderabad \textsuperscript{1},\\
        CDAC Noida \textsuperscript{2}, IIIT Bhubaneswar \textsuperscript{3}\\
        \{sankalp.bahad, pruthwik.mishra\}@research.iiit.ac.in \\
        karunesharora@cdac.in, rakesh@iiit-bh.ac.in\\
        \{dipti, param.krishna\}@iiit.ac.in \\
}

\begin{document}
\maketitle
\begin{abstract}
Named Entity Recognition (NER) is a useful component in Natural Language Processing (NLP) applications. It is used in various tasks such as Machine Translation, Summarization, Information Retrieval, and Question-Answering systems. The research on NER is centered around English and some other major languages, whereas limited attention has been given to Indian languages. We analyze the challenges and propose techniques that can be tailored for Multilingual Named Entity Recognition for Indian Languages. We present a human annotated named entity corpora of $\sim$40K sentences for 4 Indian languages from two of the major Indian language families. 
Additionally,we present a multilingual model fine-tuned on our dataset, which achieves an F1 score of $\sim$0.80 on our dataset on average. We achieve comparable performance on completely unseen benchmark datasets for Indian languages which affirms the usability of our model.
\end{abstract}

\section{Introduction}

Named entities are usually real world objects that are denoted by proper names such as ``Location", ``Person'', ``Organization", etc. Named Entity Recognition (NER) is defined as a process of classifying each named entity into a category within a given piece of text. NER is very useful in the understanding of the structure and content of the textual information, and it also plays a pivotal role in various NLP applications.


India has a wide range of languages, where each language has a unique structure, script, grammar, and other linguistic characteristics. Considering India's linguistic diversity, designing accurate and robust NERs for Indian languages bears even greater significance. 
We also encounter different challenges while working with NER in an Indian language setup, mainly Hindi, Urdu, Telugu and Odia. These challenges mainly arise due to the following reasons:
\begin{enumerate}
    \item \textbf{Absence of Fixed Word Order}: Indian languages are free word ordered languages, where words can be moved around without changing the meaning of the sentence.
    \item \textbf{Absence of Capitalization}: Indian language scripts do not have capitalization which makes it difficult to recognize the proper nouns in a sentence or phrase unlike English and other European languages.
    \item \textbf{Spelling Variations}: Many Indian languages show the property of variations in spellings of the words. 
    
    \item \textbf{Variation in Word Senses}: In Indian languages, a single word can have multiple meanings based on its sense of usage. This might lead to a case where a word might belong to two different named entities, which can only be determined based on the context.
\end{enumerate}

The emergence of models such as Bidirectional Encoder Representations from Transformers (BERT) \cite{devlin-etal-2019-bert} and many of its variants has added a new dimension to NER with the possibility of developing multilingual NER solutions. This was made possible due to the training data of these models, that consisted of multiple languages. These models, unlike traditional machine learning models, demonstrated the ability of knowledge transfer across languages. This made NER more adaptable and accessible to low resource languages, like many of the Indian languages, which are still largely unexplored and low resourced.

Many Indian languages suffer from lack of labelled data, linguistic resources, and NLP toolkits which is required for designing specific language related features for most of the machine learning models. This issue can easily be resolved by the multilingual neural models by offering a viable solution of knowledge transfer from high to low resource languages. Fine-tuning a single multilingual model can leverage the linguistic knowledge encoded with the model. We experiment with different multilingual pre-trained models and show their efficacies with a strong focus on the availability of resources. 

\section{Related Work}

The previous works in this field of NER have mainly explored the challenges and opportunities of NER techniques in multilingual settings. Researchers have developed and fine tuned some multilingual NER models, that help perform NER across multiple languages \cite{nothman2013learning}. These models rely on pre-trained transformer based architectures, for example: BERT, RoBERTa \cite{zhuang-etal-2021-robustly}, XLM-RoBERTa \cite{conneau-etal-2020-unsupervised}. It has been observed that cross lingual transfer learning is extremely useful and effective for low resource languages, where NER models pre-trained on high resource languages are adapted for low resource languages.  The research has also focused on creating and curating multilingual corpora encompassing a large range of languages, that prove to be valuable resources for training and evaluating multilingual NER models.

There has been significant amount of work regarding datasets and other resources using pre-trained transformer models. Naamapadam \cite{mhaske-etal-2023-naamapadam} and HiNER \cite{murthy2022hiner} are two widely used publicly available datasets for Indian language NER.
\begin{enumerate}
    \item Naamapadam Dataset: Naamapadam consists of data from 11 major Indian languages from two language families. The dataset contains more than 400k sentences annotated with a total of at least 100k entities from three standard entity categories (Person, Location, and, Organization) for 9 out of the 11 languages. It is a significant resource for NER in Indian Languages.
    \item HiNER Dataset: This is another NER dataset by annotating data from the ILCI tourism domain \cite{jha-2010-tdil} and a subset of the news domain corpus \cite{goldhahn-etal-2012-building} in Hindi. This dataset includes a total of 108,608 sentences and 11 tags.
\end{enumerate}

\section{Named Entity Annotation}
For the task of NER, we annotated data from two domains, general and governance. At least 2 annotators with post graduation education were involved in the task for each language.  Named entities are annotated for following 4 languages where 3 are from the Indo Aryan family and 1 from Dravidian family (shown in sequence): \textbf{Hindi}, \textbf{Odia}, \textbf{Urdu}, and \textbf{Telugu}. For Hindi, 7 annotators were included. The average inter-annotator agreement for all four languages was 0.95, which shows good agreement among the annotators. 
The agreement scores are evaluated on 200 sentences for each language. We compute Cohen's Kappa measure for this. For Hindi, we compute the average of Cohen's scores among all possible combinations of the raters. Language-wise inter-annotator agreement scores are reported in Table~\ref{tab:iaa_score}. 6 tags were chosen for named entity tagging, which are detailed in Table~\ref{tab:ne_tags} followed by the examples of Person, Location, and Organization entities in all languages.
\begin{table}[ht]
    \centering
    \begin{tabular}{c|c}
        \textbf{Language} & \textbf{Agreement Score} \\\hline
        Hindi & 0.96\\
        Odia & 0.94\\
        Telugu & 0.95\\
        Urdu & 0.96\\\hline
    \end{tabular}
    \caption{Language Wise Inter Annotator Agreement Scores}
    \label{tab:iaa_score}
\end{table}
\begin{table}[H]
    \centering
    \begin{tabular}{|c|c|c|}\hline
        \textbf{Tag} & \textbf{Desc} & \textbf{Example}\\ \hline
        NEP & Person names & Virat Kohli\\
        NEL & Locations & New Delhi\\
        NEO & Organization Names & IIT-Delhi\\
        NEAR & Artefacts & Taj Mahal\\
        NEN & Number & fifteen thousand\\
        NETI & Time and Date & 5th December\\\hline
    \end{tabular}
    \caption{Named Entity Tags}
    \label{tab:ne_tags}
\end{table}


\begin{figure}[h]
    \centering
    \includegraphics[width=1\linewidth]{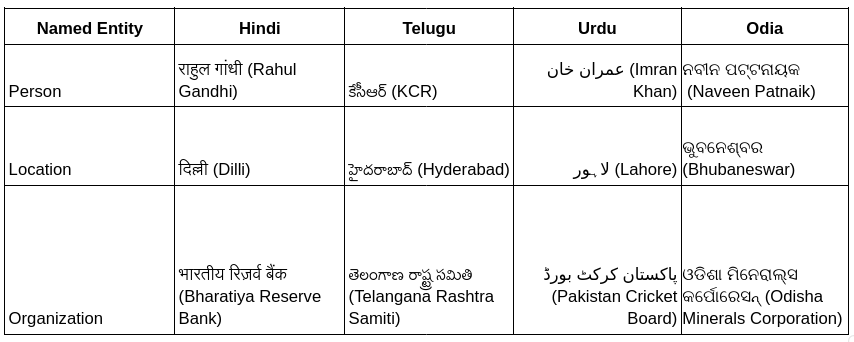}
    \caption{Enter Caption}
    \label{fig:enter-label}
\end{figure}

\section{Methodology}
We first explored various datasets and models available for Hindi Named Entity Recognition. As our named entity annotated corpus is annotated with a different tagset, we could not make use of the existing models directly. In this pursuit, we explored different fine-tuning techniques to develop a model tailor-made for our tagset.

We experiment with two approaches for the creation of monolingual models. First approach is to fine-tune a baseline BERT model for our task, and the second approach fine-tunes a BERT based NER model for our task, on our annotated dataset.
As our basic model, we select XLM-RoBERTa-Base \cite{conneau-etal-2020-unsupervised} model, which is a transformer based architecture designed for multilingual natural language understanding tasks. This model is pretrained on a vast multilingual corpus and hence is capable of efficiently handling multiple languages, which makes it well suited for the multilingual NER task. The selection of this model for multilingual NER in Indian languages can be further justified by its strong performance in various NLP tasks and its ability to generalize well across languages. Its multilingual pre-training enables it to capture linguistic nuances in different languages, including those present in Indian languages.

As our main focus had been creating a multilingual model for low resource languages, we found multiple ways of improving the results for NER for low resource languages, some of them are as follows:


\begin{itemize}
    \item One method involves extending the vocabulary, encoders, and decoders to accommodate target languages and continuing pretraining on the target language. Subsequently, pretraining continues using monolingual data in the target language.

    \item Another approach is to use alignment models like MUSE or VecMap with bilingual dictionaries to initialize the embeddings of new vocabulary, instead of randomly initializing them.

    \item An alternative strategy involves cross-lingual and progressive transfer learning, where language model training for low-resource languages begins with a large language model for a high-resource language, including overlapping vocabulary.

    \item Building extensive corpora from existing parallel data can also be beneficial. This approach enables the creation of high-quality training data for multilingual models and facilitates the training of models for low-resource languages that may lack sufficient training data.
\end{itemize}

Out of all these available methods, we find the approach that uses cross lingual and progressive transfer learning, to train language models for low resource languages with language model for high resource languages by appending the vocabulary. This method worked well for languages belonging to the same language family.



We also try taking a different approach of converting the scripts from native to roman script and carrying out the experiments on the multilingual model, but it was observed that the model trained on native scripts was performing better than the model trained on the roman scripts. A reason for this behaviour can be the absence of roman scripts for the corresponding native scripts of the language in the training data of the pretrained XLM RoBERTa \cite{conneau-etal-2020-unsupervised} base model. Hence, no further exploration was done in this direction. 

We also evaluated the dataset on the CRF \cite{10.5555/645530.655813,patil2020named} model, which as expected did not give a good result due to the fact that it was not a pre-trained model. The major limitation of a CRF model lies in it inability to transfer knowledge for reusability. Hence, we did not continue any exploration in that direction.

\section{Experiments}

 Table~\ref{tab:data_split} shows a list of languages and the corresponding number of sentences in their training, testing, and validation datasets. We have released label-wise count for all languages in the Appendix section. As a part of this work, we release annotated datasets of 4 languages with different degrees of morphological richness: Hindi, Urdu, Odia, and Telugu.

\begin{table}[H]
\centering
\begin{tabular}{|l|c|c|c|}
\hline
\textbf{Language} & \textbf{Train} & \textbf{Test} & \textbf{Dev} \\
\hline
Hindi & 11076 & 1389 & 1389 \\
Urdu & 8720 & 1096 & 1094 \\
Odia & 12109 & 1519 & 1517 \\
Telugu & 2993 & 384 & 384 \\
\hline
\end{tabular}
\caption{Language Dataset Split in terms of Sentences}
\label{tab:data_split}
\end{table}

\begin{table*}[h]
\centering
\begin{tabular}{|l|c|c|c|c|c|c|}
\hline
\textbf{Label} & \multicolumn{2}{c|}{\textbf{Dev Dataset}} & \multicolumn{2}{c|}{\textbf{Test Dataset}} \\
\hline
& \textbf{Indic NER F1-Score} & \textbf{HiNER F1-Score} & \textbf{Indic NER F1-Score} & \textbf{HiNER F1-Score} \\
\hline
NEL & 0.68 & 0.68 & 0.73 & 0.84 \\
NEO & 0.38 & 0.40 & 0.31 & 0.42 \\
NEP & 0.77 & 0.68 & 0.69 & 0.64 \\
\hline
Micro Avg & 0.60 & 0.59 & 0.55 & 0.66 \\
Macro Avg & 0.61 & 0.59 & 0.57 & 0.63 \\
Weighted Avg & 0.64 & 0.62 & 0.61 & 0.68 \\
\hline
\end{tabular}
\caption{Comparison of F1-Scores for Indic NER and HiNER models on Dev and Test Datasets}
\label{tab:merged_both_datasets}
\end{table*}

Our experiments include reviewing of the earlier methods including Conditional Random Fields and neural based named entity taggers. In this, we analyze the pre-trained models and datasets released as Indic NER model and Naamapadam dataset \cite{mhaske-etal-2023-naamapadam} and HiNER \cite{murthy2022hiner} .

Our experiments include testing Indic NER and HiNER on our annotated dataset, where we record an F1 score between 0.55 to 0.65 for the dev and test sentences of the gold dataset. We refer to our dataset as \textit{gold} dataset and this convention is used in the future tables and figures.  These experiments are conducted to visualize the performance of different models and adapting them towards developing a customized model for our gold dataset. As an initial experiment, we test the publicly available models on each other to assess their performance which are reported in Table~\ref{tab:merged_both_datasets}.

We then proceed towards creating a monolingual model for Hindi. Our hypothesis is that a model that is already trained on NER task is expected to outperform the base model with no knowledge about the NER task. We validate our hypothesis by fine-tuning a baseline BERT model (not trained for an NER task) on our annotated dataset and fine-tuning a BERT based NER model (HiNER) on our annotated dataset. This experiment is carried out on all the tags of our dataset. We report accuracies between BERT \cite{devlin-etal-2019-bert} based NER model and baseline BERT based model. As expected, the model which is a result of fine-tuning on HiNER model performs better than fine-tuning on baseline BERT model.

We then combine all the data from different languages and train a multilingual model. We experiment with changing of scripts i.e converting all the data to the same script before finetuning, to check whether the new model performs better or worse than the original model. We convert all our data to Roman script for this purpose. We then fine-tune the RoBERTa base model on Naamapadam dataset and gold dataset as the part of the comparative study between native script and roman script. 

In the fine-tuning approach used, we combine all the training data for all languages and fine-tune the monolingual model on this combined data. We then analyze the performance of each language on the multilingual model.

\section{Results and Discussion}

\subsection{Review of earlier methods}
In this section, we look at the results of the experiments we performed on the existing models. We used the metrics from the Seqeval \cite{seqeval} library to calculate F1 Scores and Classification reports. 



Table \ref{tab:merged_both_datasets} shows the performance of the IndicNER \cite{mhaske-etal-2023-naamapadam} and HiNER \cite{murthy2022hiner} models on the test and dev sets of our datasets. From the scores, we clearly observe that the model is unable to predict the NEO tags appropriately. 

Results of the test set of the data released by HiNER on IndicNER model and test set of the data released by AI4Bharat on HiNER model are shown in Tables~\ref{tab:indic-on-hiner} and \ref{tab:hiner-on-indic} respectively. 

These results show the quality of our annotated datasets and how the already available NER models perform on this dataset. Our dataset gives decent scores in zero shot tests on the IndicNER and HiNER models. Further experiments include fine-tuning these models on our dataset and analyzing their results. 

\begin{table} [htbp]
\centering
\begin{tabular}{|l|ccc|}    
\hline
\textbf{Label} & \textbf{Precision} & \textbf{Recall} & \textbf{F1-Score} \\
\hline
LOC & 0.88 & 0.65 & 0.75 \\
ORG & 0.62 & 0.59 & 0.60 \\
PER & 0.72 & 0.83 & 0.78 \\
\hline
\textbf{Micro Avg} & 0.82 & 0.67 & 0.74 \\
\textbf{Macro Avg} & 0.74 & 0.69 & 0.71 \\
\textbf{Weighted Avg} & 0.83 & 0.67 & 0.74 \\
\hline
\end{tabular}
\caption{Indic NER model on HiNER Dataset}
\label{tab:indic-on-hiner}
\end{table}

\begin{table} [htbp]
\centering
\begin{tabular}{|l|ccc|}
\hline
\textbf{Label} & \textbf{Precision} & \textbf{Recall} & \textbf{F1-Score} \\
\hline
LOC & 0.83 & 0.78 & 0.80 \\
ORG & 0.72 & 0.65 & 0.69 \\
PER & 0.86 & 0.80 & 0.83 \\
\hline
\textbf{Micro Avg} & 0.81 & 0.75 & 0.78 \\
\textbf{Macro Avg} & 0.80 & 0.74 & 0.77 \\
\textbf{Weighted Avg} & 0.81 & 0.75 & 0.78 \\
\hline
\end{tabular}
\caption{HiNER model on Naamapadam dataset}
\label{tab:hiner-on-indic}
\end{table}

\subsection{Building new models}

The test results of the baseline BERT model fine-tuned on our annotated Hindi data is shown in the Table \ref{tab:baseline-bert} and that of the HiNER model fine-tuned on our annotated Hindi data is shown in the Table \ref{tab:hiner-finetuned}. We observe close to an overall F1 score of 0.82 on the baseline BERT model for our dataset, and an overall F1 score of 0.83 on HiNER Model fine-tuned. This supports our assumption of getting a better score on model fine-tuned on an existing NER model than by fine-tuning a bare BERT model. 


\begin{table}[H]
\centering
\begin{tabular}{|l|ccc|}
\hline
\textbf{Label} & \textbf{Precision} & \textbf{Recall} & \textbf{F1-Score} \\
\hline
NEAR & 0.32 & 0.44 & 0.37 \\
NEL & 0.83 & 0.87 & 0.85 \\
NEN & 0.87 & 0.90 & 0.89 \\
NEO & 0.58 & 0.55 & 0.56 \\
NEP & 0.85 & 0.85 & 0.85 \\
NETI & 0.73 & 0.75 & 0.74 \\
\hline
\end{tabular}
\caption{Performance of the baseline BERT model on our dataset}
\label{tab:baseline-bert}
\end{table}

\begin{table}[H]
\centering
\begin{tabular}{|l|ccc|}
\hline
\textbf{Label} & \textbf{Precision} & \textbf{Recall} & \textbf{F1-Score} \\
\hline
NEAR & 0.19 & 0.28 & 0.22 \\
NEL & 0.88 & 0.92 & 0.90 \\
NEN & 0.85 & 0.89 & 0.87 \\
NEO & 0.60 & 0.57 & 0.59 \\
NEP & 0.81 & 0.85 & 0.83 \\
NETI & 0.75 & 0.80 & 0.78 \\
\hline
\end{tabular}
\caption{Performance of the HiNER model on our dataset}
\label{tab:hiner-finetuned}
\end{table}

Table \ref{tab:bert-hiner} shows the comparison between the F1 scores on the Test set, of the baseline BERT model and the HiNER model fine-tuned on our Hindi annotated data. 

\begin{table} [H]
\centering
\begin{tabular}{|c|c|}
\hline
\textbf{Model} & \textbf{F1 Score} \\
\hline
baseline BERT Model & 0.8205 \\
HiNER Model & 0.8316 \\
\hline
\end{tabular}
\caption{Comparison of F1 Scores between baseline BERT and HiNER Models}
\label{tab:bert-hiner}
\end{table}


The above results show that using an already trained NER model for fine-tuning is better than using a baseline BERT model for fine-tuning in the monolingual Hindi case. 



\begin{table}[H]
\centering
\begin{tabular}{|c|c|c|}
\hline
\textbf{Test-Dataset} & \textbf{Monolingual} & \textbf{Multilingual} \\
& & \textbf{(Combined)} \\
\hline
Gold-Hindi & 0.8205 & 0.8105 \\
Gold-Odia & 0.7546 & 0.7715 \\
Gold-Telugu & 0.7632 & 0.7555 \\
Gold-Urdu & 0.8285 & 0.8331 \\

\hline
\end{tabular}
\caption{F1 Scores for a Multilingual Model}
\label{tab:multi}
\end{table}

Table \ref{tab:multi} shows the F1 Scores of different languages on the monolingual and multilingual models for all the four languages on the Gold dataset. We observe the monolingual and multilingual scores to be in the range of 0.75 to 0.83. The multilingual models exhibit an increase in scores for Odia and Urdu, whereas there is a slight dip in the scores for Telugu and Hindi. A possible reason for this can be that Telugu and Hindi belong to different language families. Overall, multilingual models demonstrates comparable results to monolingual models, exhibiting the capability and effectiveness in multiple languages being handled simultaneously. 

We also tested our models on Naamapadam test set. The results are not very useful as that Indic-NER can only predict 3 tags, whereas our developed model predicts all the 7 tags. 



\section*{Acknowledgement}
This annotated corpora has been developed under the Bhashini project funded by Ministry of Electronics and Information Technology (MeitY), Government of India. We thank MeitY for funding this work. We sincerely thank the annotators who developed this corpora whose names are added in the appendix.
\section{Conclusion and Future Work}

We introduce a specialized NER dataset tailored for four Indian languages. Our experiments with established NER models on this dataset provide valuable insights for fine-tuning. Our proposed fine-tuning technique paves a way for NER in low resource languages. Techniques such as transfer learning and architectural modifications can further be explored to improve the model. We propose augmenting our dataset with additional annotated sentences. Adding data from other Indian languages can potentially lead to substantial performance improvements.

\bibliography{custom}
\newpage
\onecolumn
\section*{Appendix}

    
\subsection*{Data Statistics}
Figures \ref{tab:odia-label}, \ref{tab:telugu-label}, \ref{tab:hindi-label}, and \ref{tab:urdu-label} show a list of label counts for Test, Validation, and Train datasets for Odia, Telugu, Hindi, and Urdu language.
Tables \ref{tab:hindi}, \ref{tab:telugu}, \ref{tab:urdu}, \ref{tab:odia} show a comparative study of the classification reports for Hindi, Telugu, Urdu, and Odia language for the monolingual and multilingual models.

\begin{table}[htbp]
   \centering
   \begin{minipage}{0.45\textwidth}
       \centering
       \begin{tabular}{|l|c|c|c|}
           \hline
           \textbf{Label} & \textbf{Test} & \textbf{Validation} & \textbf{Train} \\
                          & \textbf{Count} & \textbf{Count} & \textbf{Count} \\
           \hline
           NEAR & 24 & 24 & 183 \\
           NEP & 59 & 59 & 471 \\
           NETI & 64 & 64 & 509 \\
           NEL & 87 & 87 & 695 \\
           NEO & 35 & 35 & 280 \\
           NEN & 8 & 8 & 60 \\
           \hline
       \end{tabular}
       \caption{Odia Data Label Split}
       \label{tab:odia-label}
   \end{minipage}
   \hfill
   \begin{minipage}{0.45\textwidth}
       \centering
       \begin{tabular}{|l|c|c|c|}
           \hline
           \textbf{Label} & \textbf{Test} & \textbf{Validation} & \textbf{Train} \\
                          & \textbf{Count} & \textbf{Count} & \textbf{Count} \\
           \hline
           NEN & 76 & 76 & 606 \\
           NETI & 17 & 17 & 130 \\
           NEP & 14 & 14 & 110 \\
           NEL & 5 & 5 & 13 \\
           NEO & 8 & 8 & 57 \\
           NEAR & 5 & 5 & 13 \\
           \hline
       \end{tabular}
       \caption{Telugu Data Label Split}
       \label{tab:telugu-label}
   \end{minipage}
\end{table}

\begin{table}[htbp]
   \centering
   \begin{minipage}{0.45\textwidth}
       \centering
       \begin{tabular}{|l|c|c|c|}
           \hline
           \textbf{Label} & \textbf{Test} & \textbf{Validation} & \textbf{Train} \\
                          & \textbf{Count} & \textbf{Count} & \textbf{Count} \\
           \hline
           NEP & 97 & 97 & 774 \\
           NETI & 154 & 154 & 1226 \\
           NEN & 295 & 295 & 2357 \\
           NEL & 93 & 93 & 742 \\
           NEO & 60 & 60 & 476 \\
           NEAR & 15 & 15 & 112 \\
           \hline
       \end{tabular}
       \caption{Hindi Data Label Split}
       \label{tab:hindi-label}
   \end{minipage}
   \hfill
   \begin{minipage}{0.45\textwidth}
       \centering
       \begin{tabular}{|l|c|c|c|}
           \hline
           \textbf{Label} & \textbf{Test} & \textbf{Validation} & \textbf{Train} \\
                          & \textbf{Count} & \textbf{Count} & \textbf{Count} \\
           \hline
           NEL & 106 & 106 & 847 \\
           NEN & 213 & 213 & 1700 \\
           NETI & 5 & 5 & 31 \\
           NEO & 16 & 16 & 126 \\
           NEP & 39 & 39 & 303 \\
           NEAR & 5 & 5 & 36 \\
           \hline
       \end{tabular}
       \caption{Urdu Data Label Split}
       \label{tab:urdu-label}
   \end{minipage}
\end{table}


\subsection*{Label Wise Results}
\begin{table}[h!]
    \centering
    \begin{tabular}{|l|c|c|c|c|c|c|}
    \hline
    \textbf{Category} & \multicolumn{3}{c|}{\textbf{Monolingual}} & \multicolumn{3}{c|}{\textbf{Multilingual}} \\
    \hline
    & \textbf{Precision} & \textbf{Recall} & \textbf{F1-score} & \textbf{Precision} & \textbf{Recall} & \textbf{F1-score} \\
    \hline
    NEAR & 0.52 & 0.58 & 0.55 & 0.52 & 0.54 & 0.53 \\
    NEL & 0.85 & 0.87 & 0.86 & 0.85 & 0.86 & 0.85 \\
    NEN & 0.94 & 0.90 & 0.92 & 0.95 & 0.91 & 0.93 \\
    NEO & 0.66 & 0.66 & 0.66 & 0.63 & 0.65 & 0.64 \\
    NEP & 0.85 & 0.84 & 0.84 & 0.82 & 0.81 & 0.82 \\
    NETI & 0.69 & 0.71 & 0.70 & 0.64 & 0.68 & 0.66 \\
    \hline
    \end{tabular}
    \caption{Comparison of Hindi Named Entity Recognition Performance in Monolingual and Multilingual Settings}
    \label{tab:hindi}
\end{table}


\begin{table}[h!]
    \centering
    \begin{tabular}{|l|c|c|c|c|c|c|}
    \hline
    \textbf{Category} & \multicolumn{3}{c|}{\textbf{Monolingual}} & \multicolumn{3}{c|}{\textbf{Multilingual}} \\
    \hline
    & \textbf{Precision} & \textbf{Recall} & \textbf{F1-score} & \textbf{Precision} & \textbf{Recall} & \textbf{F1-score} \\
    \hline
    NEAR & 0.67 & 0.50 & 0.57 & 0.75 & 0.50 & 0.60 \\
    NEL & 0.70 & 0.58 & 0.64 & 0.80 & 0.57 & 0.67 \\
    NEN & 0.87 & 0.90 & 0.88 & 0.84 & 0.91 & 0.87 \\
    NEO & 0.42 & 0.56 & 0.48 & 0.50 & 0.56 & 0.53 \\
    NEP & 0.59 & 0.57 & 0.58 & 0.58 & 0.70 & 0.64 \\
    NETI & 0.49 & 0.74 & 0.59 & 0.43 & 0.52 & 0.47 \\
    \hline
    \end{tabular}
    \caption{Comparison of Telugu Named Entity Recognition Performance in Monolingual and Multilingual Settings}
    \label{tab:telugu}
\end{table}

\begin{table}[h!]
    \centering
    \begin{tabular}{|l|c|c|c|c|c|c|}
    \hline
    \textbf{Category} & \multicolumn{3}{c|}{\textbf{Monolingual}} & \multicolumn{3}{c|}{\textbf{Multilingual}} \\
    \hline
    & \textbf{Precision} & \textbf{Recall} & \textbf{F1-score} & \textbf{Precision} & \textbf{Recall} & \textbf{F1-score} \\
    \hline
    NEAR & 0.33 & 0.20 & 0.25 & 0.50 & 0.40 & 0.44 \\
    NEL & 0.82 & 0.80 & 0.81 & 0.78 & 0.76 & 0.77 \\
    NEN & 0.96 & 0.90 & 0.93 & 0.98 & 0.90 & 0.94 \\
    NEO & 0.39 & 0.37 & 0.38 & 0.49 & 0.47 & 0.48 \\
    NEP & 0.77 & 0.64 & 0.70 & 0.84 & 0.62 & 0.71 \\
    NETI & 0.58 & 0.78 & 0.67 & 0.67 & 0.89 & 0.76 \\
    \hline
    \end{tabular}
    \caption{Comparison of Urdu Named Entity Recognition Performance in Monolingual and Multilingual Settings}
    \label{tab:urdu}
\end{table}

\begin{table}[h!]
    \centering
    \begin{tabular}{|l|c|c|c|c|c|c|}
    \hline
    \textbf{Category} & \multicolumn{3}{c|}{\textbf{Monolingual}} & \multicolumn{3}{c|}{\textbf{Multilingual}} \\
    \hline
    & \textbf{Precision} & \textbf{Recall} & \textbf{F1-score} & \textbf{Precision} & \textbf{Recall} & \textbf{F1-score} \\
    \hline
    NEAR & 0.73 & 0.58 & 0.64 & 0.86 & 0.58 & 0.69 \\
    NEL & 0.89 & 0.82 & 0.85 & 0.90 & 0.84 & 0.87 \\
    NEN & 0.46 & 0.29 & 0.35 & 0.44 & 0.38 & 0.41 \\
    NEO & 0.65 & 0.76 & 0.70 & 0.64 & 0.70 & 0.67 \\
    NEP & 0.85 & 0.83 & 0.84 & 0.88 & 0.85 & 0.86 \\
    NETI & 0.59 & 0.70 & 0.64 & 0.66 & 0.71 & 0.68 \\
    \hline
    \end{tabular}
    \caption{Comparison of Odia Named Entity Recognition Performance in Monolingual and Multilingual Settings}
    \label{tab:odia}
\end{table}
\newpage
\subsection*{Details of Annotators}
\begin{table}[h!]
    \begin{tabular}{llll}
        \textbf{Language} & \textbf{Language Expert} & \textbf{Designation} & \textbf{Affiliation}\\\hline
        Hindi & Alpana Agarwal& Senior Language Editor & IIIT-Hyderabad\\
        & Preeti Pradhan & Senior Language Editor& IIIT-Hyderabad\\
        & Nandini Upasani & Senior Language Editor& IIIT-Hyderabad\\
        & Naresh Bansal & Senior Language Editor& IIIT-Hyderabad\\
        & Vaibhavi Kailash Kothadi &Senior Language Editor& IIIT-Hyderabad\\
        & Pranjali Kanade & Language Editor& IIIT-Hyderabad\\
        & Kaberi Sau & Senior Language Editor& IIIT-Hyderabad\\\hline
        Odia & Prakash Kumar Bhuyan & Linguist & CDAC-Noida\\
        & Bigyan Ranjan Das & Project Assistant & IIIT-Bhubaneswar\\\hline
        Telugu & Koustubha NS & Senior Language Editor & IIIT-Hyderabad\\
        & Sarala Sree Ramancharla & Senior Language Editor& IIIT-Hyderabad\\\hline
        Urdu & Mohammed Younus & Language Editor& IIIT-Hyderabad\\
        & Mohd. Noman Ali & Language Editor& IIIT-Hyderabad\\
        
    \end{tabular}
\end{table}
\end{document}